\pdfoutput=1

\documentclass[11pt]{article}

\usepackage[final]{acl}

\usepackage{times}
\usepackage{latexsym}

\usepackage{graphicx}
\usepackage{booktabs}
\usepackage{listings}
\usepackage{fontawesome5}
\usepackage{xcolor}
\usepackage{tcolorbox}
\usepackage{subcaption}

\definecolor{code-border}{rgb}{0.3,0.3,0.3}
\definecolor{code-background}{rgb}{0.975,0.975,0.975}
\definecolor{code-values}{rgb}{0.380, 0.569, 0.322}
\definecolor{code-comments}{rgb}{0.6,0.6,0.6}
\definecolor{code-classes}{rgb}{0.259,0.498,0.722}
\definecolor{code-keywords}{rgb}{0.840, 0.502, 0.781}
\definecolor{code-args}{rgb}{0.941, 0.698, 0.298}

\newcommand{\variationist}{\textsc{Variationist}}
\newcommand{\icon}{\raisebox{-1pt}{\includegraphics[width=1.2em]{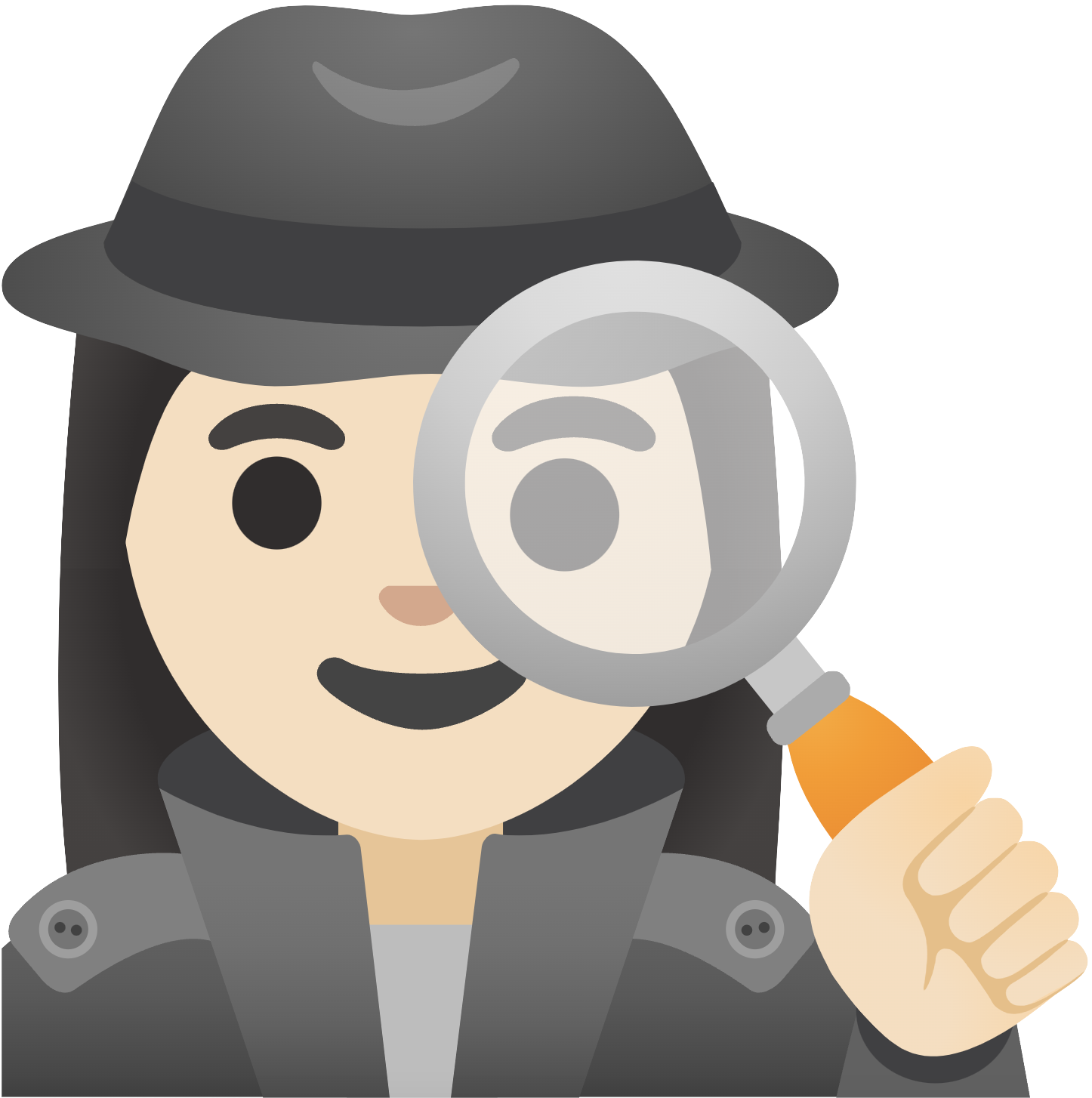}}}
\newcommand{\hficon}{\raisebox{-2pt}{\includegraphics[width=1.2em]{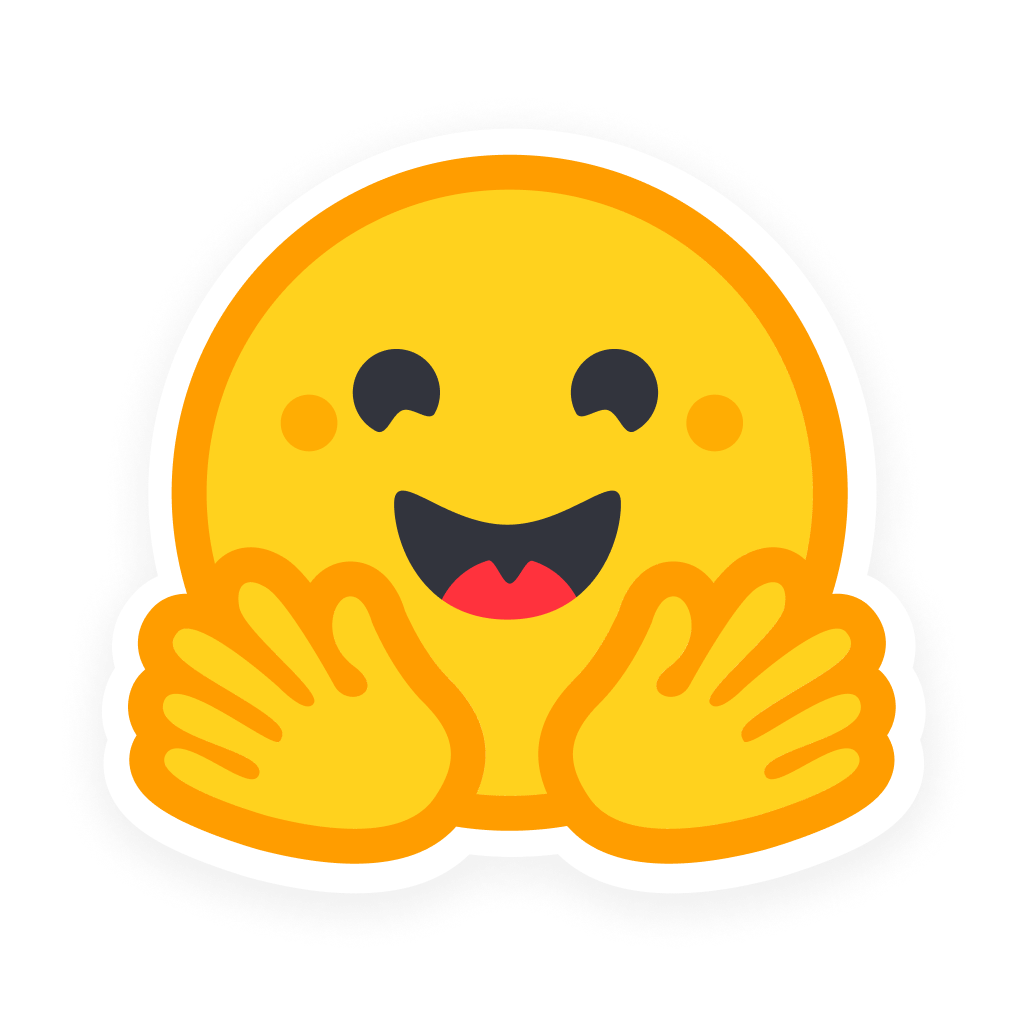}}}

\usepackage[T1]{fontenc}

\usepackage[utf8]{inputenc}

\usepackage{microtype}

\usepackage{inconsolata}

\usepackage{graphicx}

\title{\icon~\variationist:\\Exploring Multifaceted Variation and Bias in Written Language Data}

\author{Alan Ramponi,\textsuperscript{\faSun}$^*$ Camilla Casula,\textsuperscript{\faSun}\textsuperscript{\faMoon}$^*$ Stefano Menini\textsuperscript{\faSun}\\ 
\texttt{alramponi@fbk.eu}, \texttt{ccasula@fbk.eu}, \texttt{menini@fbk.eu} \\ \\
\textsuperscript{\faSun} Fondazione Bruno Kessler, Italy \\
\textsuperscript{\faMoon} University of Trento, Italy}

\begin{document}
\maketitle
\begin{abstract}
Exploring and understanding language data is a fundamental stage in all areas dealing with human language. 
It allows NLP practitioners to uncover quality concerns and harmful biases in data before training, and helps linguists and social scientists to gain insight into language use and human behavior.
Yet, there is currently a lack of a unified, customizable tool to seamlessly inspect and visualize language variation and bias across multiple \emph{variables}, language \emph{units}, and diverse \emph{metrics} that go beyond descriptive statistics.
In this paper, we introduce \variationist, a highly-modular, extensible, and task-agnostic tool that fills this gap. \variationist~handles at once a potentially unlimited combination of variable \emph{types} and \emph{semantics} across diversity and association metrics with regards to the language unit of choice, and orchestrates the creation of up to five-dimensional interactive charts for over 30 variable type--semantics combinations.
Through our case studies on computational dialectology, human label variation, and text generation, we show how \variationist~enables researchers from different disciplines to effortlessly answer specific research questions or unveil undesired associations in language data. A Python library, code, documentation, and tutorials are made publicly available to the research community.
\end{abstract}

\def\thefootnote{*}\footnotetext{These authors contributed equally to this work.}\def\thefootnote{\arabic{footnote}}

\section{Introduction}
Language data is at the core of a large body of work in many research fields and at their intersections. Language data is used to train large language models (LLMs) by natural language processing (NLP) practitioners, but also by linguists and social scientists to analyze human language and behavior. 

With a tendency in the NLP community to overlook what actually \textit{is} in the training data of models~\citep{bender-etal-2021-dangers}, especially at the level of textual information, and how different characteristics of the data can be intertwined, we propose a tool that can help in inspecting language data in a straightforward and highly customizable manner. 

\begin{figure}[t]
\centering
     \includegraphics[width=\linewidth]{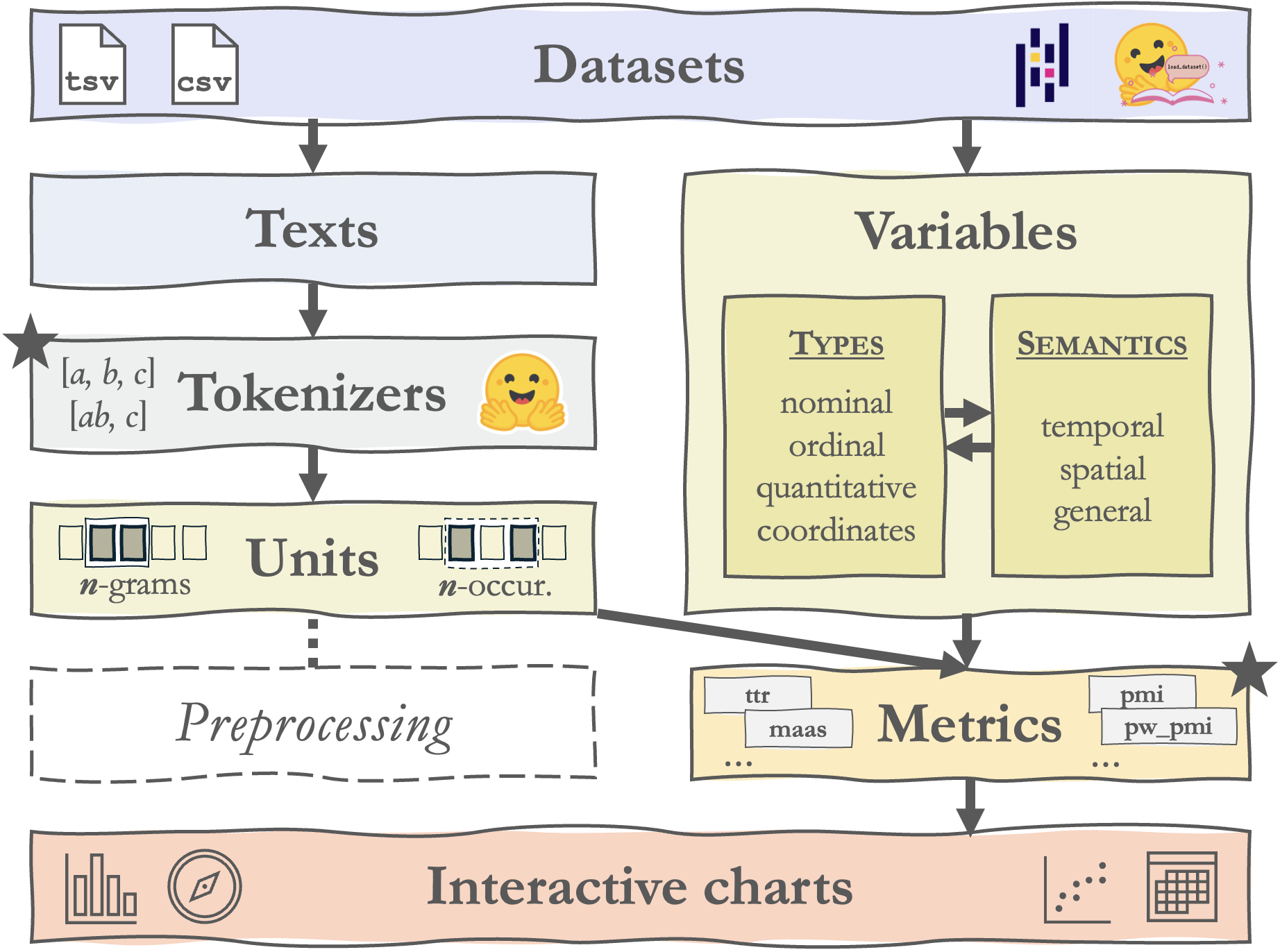}
     \caption{A high-level overview of the core elements and functionalities of \variationist. The tool computes association metrics between any unit in language and potentially unlimited variable type--semantics combinations, orchestrates the creation of interactive charts, and also supports user-defined custom components (\faStar).}
     \label{fig:overview}
\end{figure}

While some language data exploration tools already exist, especially English-centric corpus linguistics tools~\citep{anthony-2013-critical}, these cannot typically handle different types of textual \emph{units} (e.g., tokens, bigrams, characters, and more) and multiple \emph{variables} or combinations thereof, only offering surface-level \emph{metrics} that are not easily customizable, and providing low-dimensional visualization. On the other hand, modern analysis tools in NLP mainly focus on interpreting model outputs~\citep[][\emph{inter alia}]{sarti-etal-2023-inseq,attanasio-etal-2023-ferret} rather than exploring the language data in itself. 

\variationist~aims to fill this gap, offering the chance to researchers from diverse disciplines to easily explore the intersections between variables in textual corpora in a plethora of different configurations in a unified manner (cf.~Figure~\ref{fig:overview}). Additionally, \variationist~allows users to plug in their own custom tokenization functions and metrics in a seamless way, opening up an unlimited number of analysis configurations in just a few lines of code, and going beyond English-centric assumptions on what the definition of a unit in language actually is.

We demonstrate the flexibility of \variationist~through a set of case studies spanning research questions pertaining to diverse areas of research: computational dialectology, human-label variation~\citep{plank-2022-problem}, and text generation.

\paragraph{Contributions} We propose \variationist, a highly flexible and customizable tool for allowing researchers from many fields to seamlessly explore and visualize language variation and bias in textual corpora across many dimensions. We release our code,\footnote{\faGithub\::~\url{https://github.com/dhfbk/variationist}.} a \texttt{Python} library,\footnote{\faPython\,\,:~\url{https://pypi.org/project/variationist}.} a detailed documentation,\footnote{\faBook\,\,:~\url{https://variationist.readthedocs.io}.} a video presentation,\footnote{\faFilm\::~The video is available in our GitHub repository.} and a set of tutorials.\footnote{\faFlask\,\,:\:\url{https://github.com/dhfbk/variationist/tree/main/examples}.}

\section{Tool Design}

In this section, we present the overall design and aim of \variationist. In Section~\ref{sec:design-principles} we detail the guiding design principles, whereas in Section~\ref{sec:core-elements} we summarize the core elements and functionalities around which \variationist~is designed.

\subsection{Design Principles} \label{sec:design-principles}

The guiding design principles of \variationist~are summarized in the following:

\begin{itemize}
    \item \textbf{Ease of use}: \variationist~is crafted to be as accessible and customizable as possible, to serve researchers from a wide range of fields who are interested in exploring textual data;
    \item \textbf{Modularity}: \variationist~is built out of small building blocks, allowing users to pick and choose their desired features and metrics without running unnecessary calculations;
    \item \textbf{Extensibility}: \variationist~is designed to be easily extended. By virtue of its intrinsic modularity, it is conceived to let users select their preferred features, and import their own custom tokenizers and metrics into the tool.
\end{itemize}

\subsection{Core Elements and Functionalities} \label{sec:core-elements}

\variationist~is designed around a set of core elements useful for computation and visualization. We provide details on each of them in the following.

\paragraph{\textsc{Datasets}} The main input for the analysis. Datasets can be provided in the form of \emph{i)} tab-separated (\texttt{tsv}) or \emph{ii)} comma-separated (\texttt{csv}) files, or \emph{iii)} pre-computed \texttt{pandas} dataframes. Moreover, \emph{iv)} any dataset from the \hficon~Hugging Face Datasets~\citep{lhoest-etal-2021-datasets} repository can be directly imported for analysis and visualization, too.

\paragraph{\textsc{Texts}} The subset of the input data, in the form of column names or indices, containing textual data. While in most scenarios only a single text column is needed, \variationist~handles up to two columns at once in the analysis. This is especially useful for exploring similarities and differences between texts associated to the same labels and/or metadata.

\paragraph{\textsc{Units}} The language unit of interest, which can be anything from characters to ``words'' (whatever their definition may be) and longer sequences. \variationist~seamlessly supports $n$-grams (i.e., $n$ contiguous language units) and co-occurrences of $n$ units (not necessarily contiguous) that fall within a user-defined window size, with optional duplicate handling. For creating units, we rely on either built-in, publicly available, or user-defined tokenizers (see below). Units may optionally undergo preprocessing with lowercasing and stopword removal. In the latter case, the user can rely on off-the-shelf stopword lists across multiple languages from the \texttt{stopwords-iso}\footnote{\url{https://github.com/stopwords-iso}.} package, provide their own lists directly or as files, or combine them.

\paragraph{\textsc{Tokenizers}} Since the driver for the computation is a language unit, we need ways to segment texts into desired units. \variationist~allows the user to leverage \emph{i)} a default whitespace tokenizer that goes beyond Latin characters, \emph{ii)} any tokenizer from \hficon~Hugging Face Tokenizers~\citep{wolf-etal-2020-transformers}, or \emph{iii)} a custom tokenizer. This way we avoid any assumptions on what actually \emph{is} a language unit, also broaden the applicability of \variationist~to a wide range of language varieties.

\paragraph{\textsc{Variables}} Variables are essential components for computing association metrics with language units. While variables in NLP typically translate to human-annotated ``labels'', those may be naturally generalized to any kind of meta-information associated to textual data (e.g., genres, dates, spatial information, sociodemographic characteristics of annotators or authors). \variationist~natively supports a potentially unlimited number of variable combinations during analysis. Due to the variety of data types and semantic meanings that variables may take, each variable (i.e., column name) is defined through the following two attributes:
\begin{itemize}
    \item \textbf{Variable \emph{types}}: the type of the variable for representation purposes. It can be either \emph{nominal} (i.e., categorical variable without an intrinsic ordering/ranking), \emph{ordinal} (variable that can be ordered/ranked), \emph{quantitative} (numerical variable -- either discrete or continuous -- which may take any value), or \emph{coordinate} (position of a point on the Earth's surface, i.e., latitude or longitude);
    \item \textbf{Variable \emph{semantics}}: how the variable must be interpreted for visualization purposes. It may be either \emph{temporal} (e.g., variables such as date or time), \emph{spatial} (e.g., \emph{coordinate} variables or \emph{nominal} variables with spatial semantics such as countries, states, or provinces), or \emph{general} (any variable that does not fall in the aforementioned categories).
\end{itemize}

\paragraph{\textsc{Metrics}} The methods used for measuring associations between language units and a potentially unlimited combination of variables. \variationist~includes metrics such as pointwise mutual information~\citep[PMI;][]{fano-1961-transmission}, its positive, normalized, and weighted variants, as well as their combinations, for a total of 8 different PMI flavors. It also includes a normalized class relevance metric based on~\citet{ramponi-tonelli-2022-features} in its positive, weighted, and positive weighted versions. Besides unit--variables association metrics, \variationist~also includes lexical diversity measures such as type-token ratio~\citep[TTR;][]{johnson-1944-studies}, root TTR~\citep{guiraud-1960-problemes}, log TTR~\citep{herdan-1960-type}, and Maas' index~\citep{maas-1972-zusammenhang}. Basic statistics such as frequencies, number of texts, number of language units, duplicate instances, average text length, and vocabulary size are also provided. Finally, custom metrics can be easily defined by the user and used for subsequent analysis, therefore extending \variationist's capabilities to specific use cases.

\paragraph{\textsc{Charts}} The visual components of the tool. \variationist~orchestrates the automatic creation of interactive charts for each metric based on the combination of variable types and semantics from a previous analysis. It defines the optimal dimension or channel (e.g., \texttt{x}, \texttt{y}, \texttt{color}, \texttt{size}, \texttt{lat}, \texttt{lon}, or a dropdown component) for each variable, creating charts with up to five dimensions (of which one is reserved for the \emph{quantitative} metric score, and the other to the \emph{nominal} language unit). Possible charts currently include temporal line charts, choropleth maps, geographic and standard scatter plots, heatmaps, binned maps, and bar charts. For each metric, one or more charts are created (e.g., in the case of \emph{nominal} variable types with \emph{spatial} semantics, both a bar chart and a geographic scatter plot are created). Charts can be interactively filtered by language unit through a search input field supporting regular expressions or a dropdown menu\footnote{The choice depends on the chart type and its number of dimensions, with the goal of keeping the overall user experience and filtering time as smooth as possible.} to smoothly explore associations between units and the variables of interest.

\section{Implementation and Usage} \label{sec:implementation}

In this section, we present implementation details (Section~\ref{sec:user-facing} and~Section~\ref{sec:data-interchange}) and an example usage of our \texttt{Python} library (Section~\ref{sec:example-usage}).

\begin{figure*}[!t]
\centering
\hspace*{-0.325cm}
\resizebox{0.925\linewidth}{!}{%
\begin{lstlisting}[language=Python]
from variationist import Inspector, InspectorArgs, Visualizer, VisualizerArgs

# 1) Define the inspector arguments
ins_args = InspectorArgs(text_names=["text"], var_names=["label"], 
    metrics=["npw_pmi"], n_tokens=1, language="en", stopwords=True, lowercase=True)
    
# 2) Run the inspector and get the results
res = Inspector(dataset="data.tsv", args=ins_args).inspect()

# 3) Define the visualizer arguments
vis_args = VisualizerArgs(output_folder="charts", output_formats=["html"])

# 4) Create interactive charts for all metrics
charts = Visualizer(input_json=res, args=vis_args).create()
\end{lstlisting}
}%
\caption{Example showcasing the four steps for inspecting data and visualizing results using \variationist.}
\label{fig:usage}
\end{figure*}

\subsection{User-facing Classes} \label{sec:user-facing}

There are two main elements a typical user interacts with: \texttt{Inspector} and \texttt{Visualizer}, as well as their respective \texttt{InspectorArgs} and \texttt{VisualizerArgs}, which store all of the parameters they work with.

\paragraph{\texttt{Inspector}} The \texttt{Inspector} class takes care of orchestrating the analysis, from importing and tokenizing the data to handling variable combinations and importing and calculating the metrics. It returns a dictionary (or a \texttt{.json} file, cf.~Section~\ref{sec:data-interchange}) with all the calculated metrics for each unit of language, variable, and combination thereof, according to a set of parameters that are set by the user through the \texttt{InspectorArgs}.

\paragraph{\texttt{InspectorArgs}} Through the \texttt{InspectorArgs} class we tell \texttt{Inspector} how to carry out the analysis. While we refer the reader to our library and related resources for the full documentation (Appendix~\ref{sec:resources}), some of the analysis details that can be set using \texttt{InspectorArgs} include what texts and variable(s) of the data to focus on, whether to use $n$-grams or $n$ co-occurrences (and if so, for what values of $n$), what tokenizer to use, including any custom ones, the selection of metrics we want to calculate, whether and how to bin the variables, and more. In short, any preference regarding the analysis will have to go through \texttt{InspectorArgs}.

\paragraph{\texttt{Visualizer}} The \texttt{Visualizer} class takes care of orchestrating the creation of a variety of interactive charts for each metric and variable combination associated to the language units of interest. It leverages the results and metadata from the dictionary (or \texttt{.json} file) resulting from a prior analysis using \texttt{Inspector}, creating charts up to five dimensions using the \texttt{altair} library~\citep{vanderplas-2018-altair}.\footnote{Due to the modular design of \variationist, we aim to integrate additional visualization libraries in future releases.} It relies on \texttt{VisualizerArgs}, a class storing specific user-defined arguments for visualization.

\paragraph{\texttt{VisualizerArgs}} The \texttt{VisualizerArgs} class provides ways to customize the creation of charts and their serialization. In particular, it allows the user to specify whether to pre-filter the visualization based on selected language units (provided as lists) or top-scoring ones (by specifying a maximum per-variable amount), provide a \texttt{shapefile} for setting the background of spatial charts, and decide whether the charts have to be saved as files and in which format, among others.

\subsection{Data Interchange} \label{sec:data-interchange}

The results of an \texttt{Inspector} analysis are either \emph{i)} stored in a variable as a dictionary, or \emph{ii)} serialized in a \texttt{.json} file. While the first case comes handy for direct use by the \texttt{Visualizer} in most cases, the second option is especially useful when dealing with large datasets and a high number of variable combinations (and possible values). Indeed, serialization will enable the results to be easily used for visualization in a later stage. Details on the structure of the interchange file are in our repository.

\subsection{Example Usage} \label{sec:example-usage}

Figure~\ref{fig:usage} shows a basic usage example of \variationist, which consists of four steps: \emph{i)} defining the \texttt{InspectorArgs}, \emph{ii)} instantiating and running the computation with \texttt{Inspector}, \emph{iii)} defining the \texttt{VisualizerArgs}, and finally \emph{iv)} creating interactive charts for the previously specified metrics through the \texttt{Visualizer}. For details on all the available parameters and hands-on tutorials, we refer the reader to our resources (Appendix~\ref{sec:resources}).

\section{Case Studies}

In the following, we scratch the surface of \variationist's capabilities by presenting case studies on diverse topics, from computational dialectology (Section~\ref{sec:case-study-cd}) to human label variation (Section~\ref{sec:case-study-hlv}) and text generation (Section~\ref{sec:case-study-tg}). Three personas with different backgrounds and aims exemplify our case studies: \emph{Alice}, \emph{Bob}, and \emph{Carol}. We then provide ideas for further applications (Section~\ref{sec:food-for-thought}). Code for case studies is available in our repository.

\begin{figure*}[t]
\centering
\begin{subfigure}{.325\textwidth}
  \centering
  \includegraphics[width=.975\linewidth]{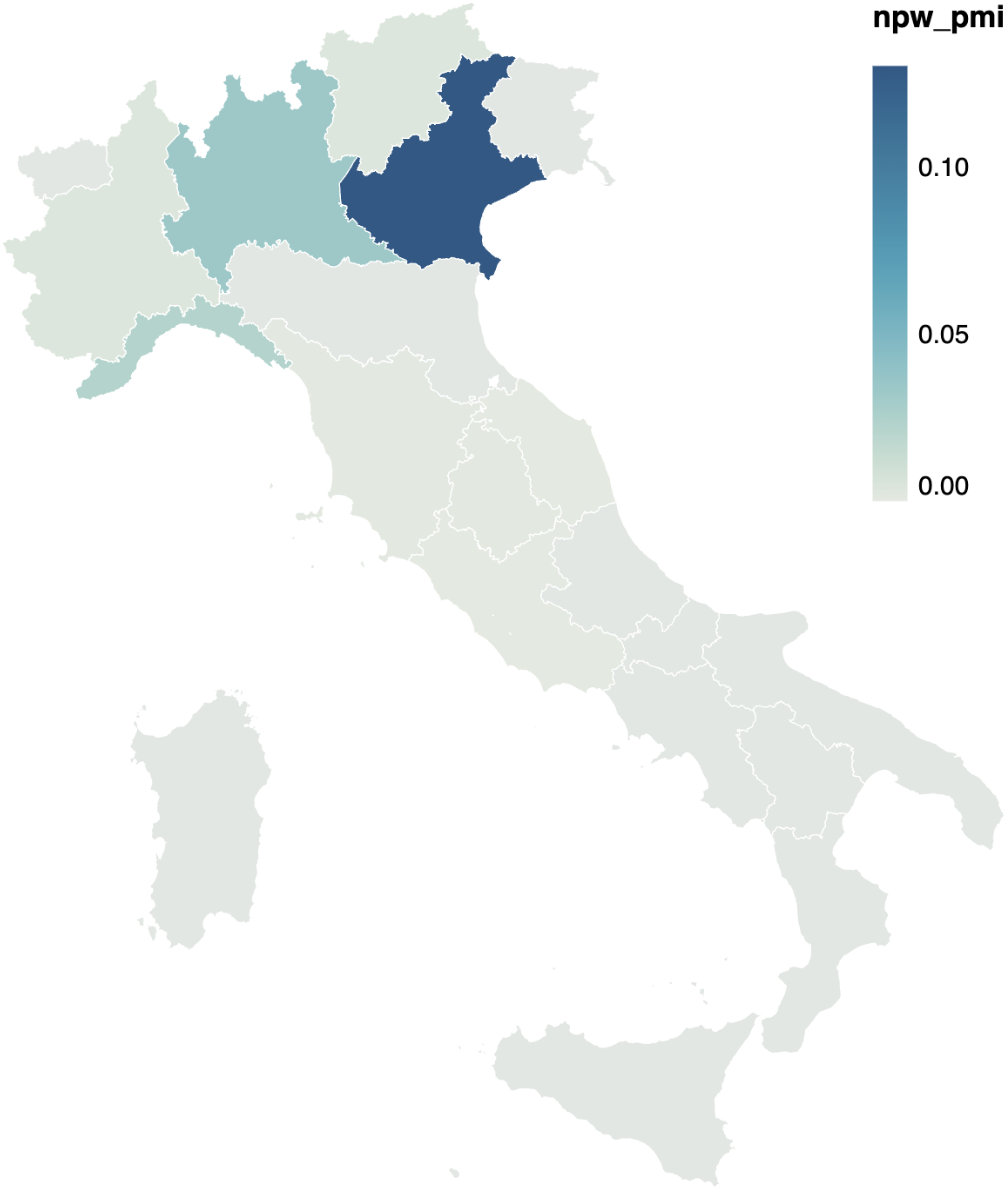}
  \caption{Choropleth map (\emph{regions}).}
  \label{fig:cs1a}
\end{subfigure} \hfill
\begin{subfigure}{.325\textwidth}
  \centering
  \includegraphics[width=.975\linewidth]{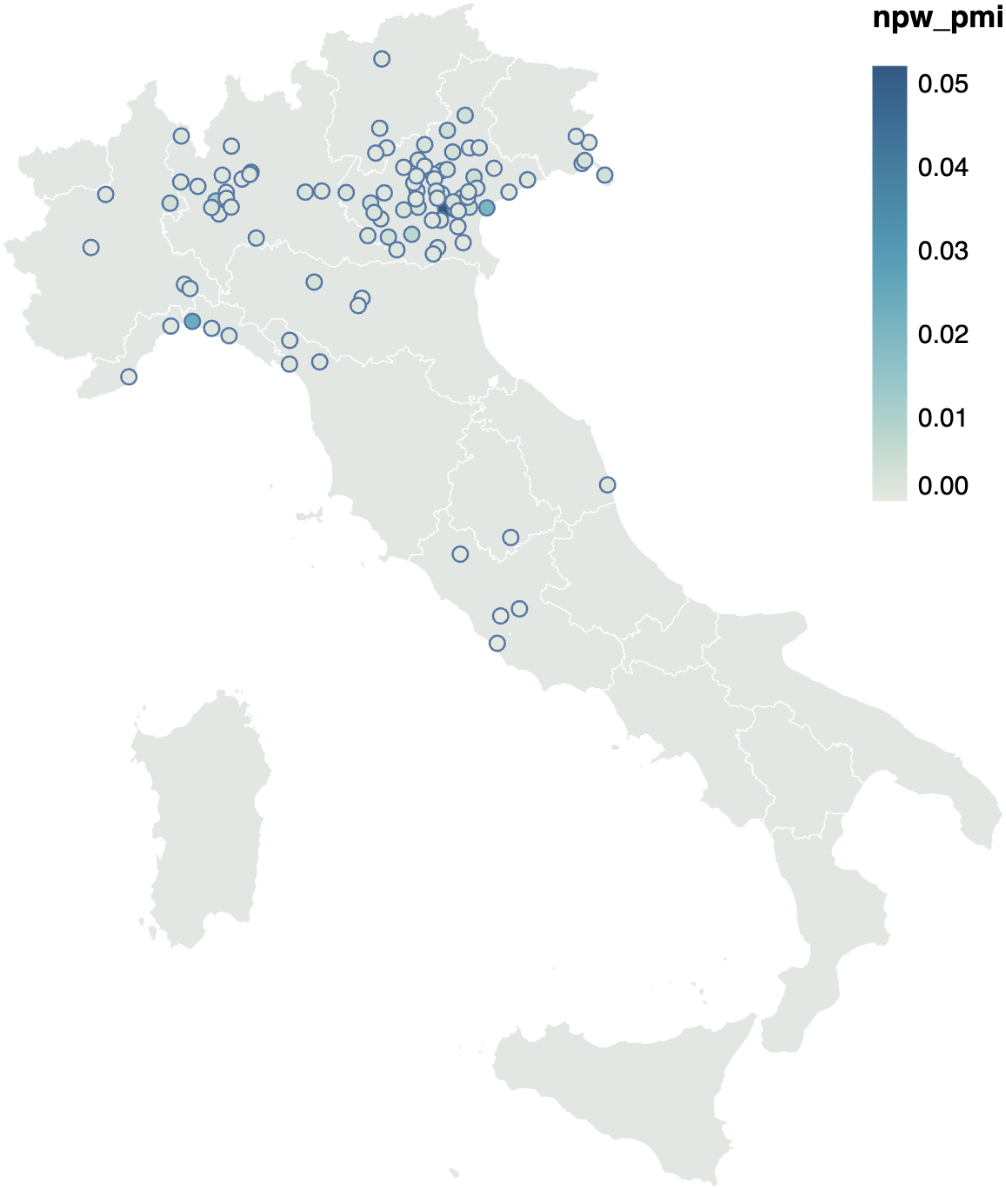}
  \caption{Geo scatter plot (\emph{municipalities}).}
  \label{fig:cs1b}
\end{subfigure} \hfill
\begin{subfigure}{.325\textwidth}
  \centering
  \includegraphics[width=.975\linewidth]{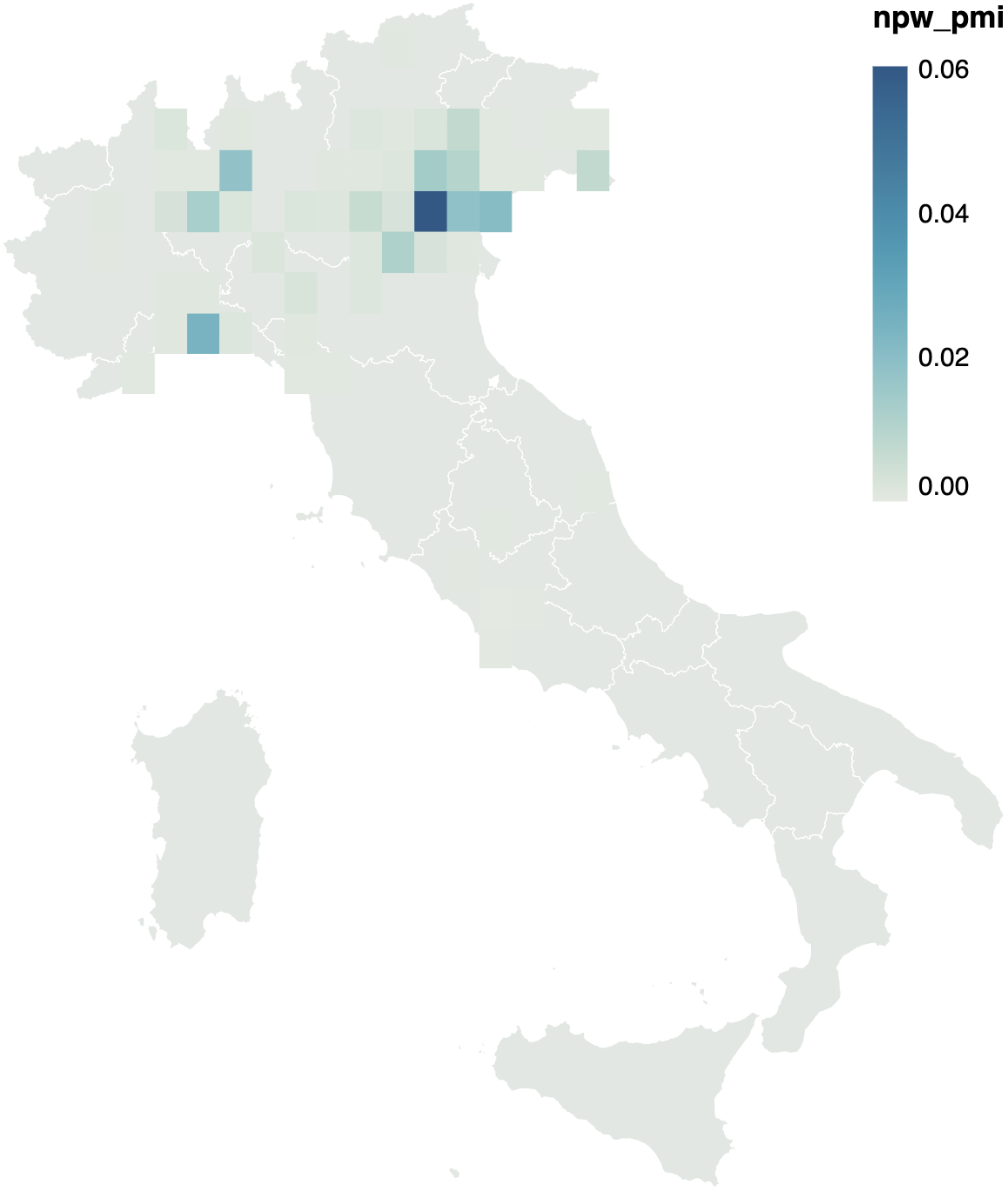}
  \caption{Binned map (\emph{custom areas}).}
  \label{fig:cs1c}
\end{subfigure}
\caption{Example visualizations for the computational dialectology case study. All the charts have been filtered to show the use of the lexical item ``\underline{ghe}'' across space within Italy at different granularities in terms of \texttt{npw\_pmi} score.}
\label{fig:cs1}
\end{figure*}

\subsection{Exploring Language Variation Across Space} \label{sec:case-study-cd}

\begin{tcolorbox}[colback=black!2!white,colframe=black!17!white,arc=0mm,fonttitle=\bfseries,coltitle=black,title=\icon~~Computational dialectology]
    \emph{Alice is a linguist interested in how language varies across space. Specifically, her research focuses on language varieties of Italy and their use in social media. Her goal is to understand in which areas selected lexical items are predominantly used.}
\end{tcolorbox}

\noindent Alice uses \textsc{DiatopIt}~\citep{ramponi-casula-2023-diatopit}, a corpus of geolocated social media posts in Italy focused on local language varieties, and provides it as input to \variationist. She specifies the \texttt{text} column of the dataset as the textual data and the \texttt{region} column as the variable of interest (setting it with \emph{nominal} type and \emph{spatial} semantics). She selects the normalized, positive, and weighted variant of PMI (\texttt{npw\_pmi}) as the metric, and chooses unigrams, derived via the default whitespace tokenizer, as the language unit. Lastly, she sets all text characters to lowercase and specifies stopword removal using a default lexicon in Italian, and extends it by providing extra unigrams to remove (i.e., the ``user'' and ``url'' placeholders). She then interactively explores the results to understand \emph{where} the lexical item ``ghe'' is predominantly used. 

As shown in the choropleth map in Figure~\ref{fig:cs1a}, the lexical item appears to be mostly used in specific regions in northern Italy, especially those where Venetian, Ligurian, and Lombard varieties are spoken. This is due to its role as an adverb and pronoun in these Romance varieties. However, language varieties of Italy cross administrative borders and multiple varieties are spoken within each region~\citep{ramponi-2024-language}.
By running \variationist~again and specifying the
\texttt{latitude} and \texttt{longitude} variables instead (both with \emph{coordinate} type and \emph{spatial} semantics), Alice gets a fine-grained picture of the actual use of the word (Figure~\ref{fig:cs1b}). Moreover, by defining \texttt{30} equally-sized intervals for the \texttt{latitude} and \texttt{longitude} variables, she obtains a binned map (Figure~\ref{fig:cs1c}) that allows her to explore the use of ``ghe'' at an intermediate granularity.

In the future, Alice would like to investigate if the use of certain lexical items underwent change over time, as a mean to assess the vitality of language varieties. By providing an additional temporal variable, she may answer her question.

\subsection{Investigating Human Subjectivity in Hate Speech Annotation} \label{sec:case-study-hlv}

\begin{tcolorbox}[colback=black!2!white,colframe=black!17!white,arc=0mm,fonttitle=\bfseries,coltitle=black,title=\icon~~Human label variation]
    \emph{Bob is a computational social scientist interested in how people perceive hateful language online. Specifically, he is interested in understanding whether annotators with different sociodemographic characteristics place greater importance to certain lexical items in determining if a message is hateful.}
\end{tcolorbox}

\noindent Bob employs the Measuring Hate Speech~\citep[MHS;][]{sachdeva-etal-2022-measuring} corpus for answering his questions. Each post in the dataset is labeled as hate speech or not and includes the demographic attributes of annotators. Bob loads the dataset from \hficon~Hugging Face Datasets\footnote{\url{https://huggingface.co/datasets/ucberkeley-dlab/measuring-hate-speech}.} and filters it to keep hateful messages only (i.e., \texttt{hatespeech=2}). For facilitating the analysis, he combines boolean columns pertaining to the same variables (e.g., \texttt{annotator\_race\_\{asian|black|}...\texttt{\}}) into single string columns (e.g., \texttt{annotator\_race} with possible values among \{\emph{asian}\texttt{|}\emph{black}\texttt{|}...\}). Then, he uses \variationist~and specifies \texttt{text} as the column containing the textual data and \texttt{npw\_relevance} as the metric for the analysis, he sets the conversion of texts to lowercase to reduce data sparsity and the removal of stopwords in English. Bob leaves the remaining parameters with default values (e.g., unigrams as units, whitespace tokenizer). As the variables of interest, he specifies \texttt{hatespeech} and either \texttt{annotator\_sexuality} or \texttt{annotator\_race} (all with \emph{nominal} type and \emph{general} semantics).

\begin{figure}[t]
\centering
\begin{subfigure}{.49\textwidth}
  \centering
  \includegraphics[width=.975\linewidth]{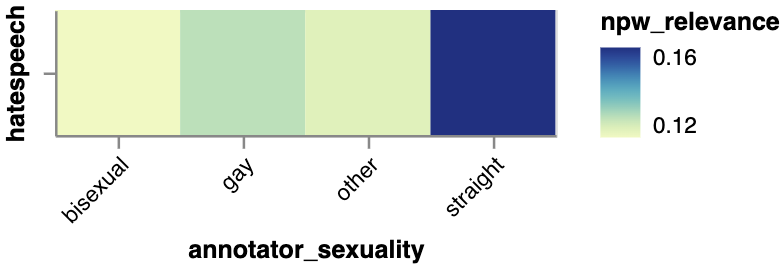}
  \caption{Heatmap for ``\underline{gay}'' across annotators' sexual orientation.}
  \label{fig:cs2a}
\end{subfigure} \par\bigskip
\begin{subfigure}{.49\textwidth}
  \centering
  \includegraphics[width=.975\linewidth]{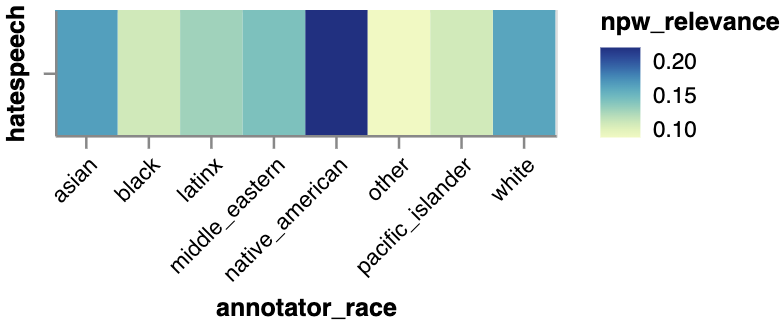}
  \caption{Heatmap for ``\underline{n*ggas}'' across annotators' race.}
  \label{fig:cs2b}
\end{subfigure}
\caption{Example visualizations for the human label variation case study. All the charts show the \texttt{npw\_relevance} score for the hateful class of specific lexical items across sociodemographics of annotators.}
\label{fig:cs2}
\end{figure}

When exploring the relationship between annotators' sexual orientation and the labels they assign to posts, Bob discovers that the lexical item ``gay'' is more indicative of the hateful class for annotators who identify as \emph{straight} compared to those who identify as \emph{bisexual}, \emph{gay}, or \emph{other} (Figure~\ref{fig:cs2a}). This may indicate that non-\emph{straight} annotators are more sensible to the nuances in the use of the term and that \emph{straight} annotators may instead occasionally use it as shortcut for hate speech annotation. Bob gets a similar finding when investigating the association of reclaimed words such as ``n*ggas'' to hateful posts across self-reported annotators' race (Figure~\ref{fig:cs2b}). The term is less associated to posts labeled as hateful by annotators who identify themselves as \emph{black} people (in-group members) compared to those annotated as hateful by most out-group members (e.g., \emph{asian}, \emph{native american}, \emph{white} people).

In summary, different lexical items may be (more or less) informative for certain labels (e.g., hate speech) depending on the sociodemographics of annotators. \variationist~can aid in speeding up the exploration of undesired associations across a combination of attributes in language data.

\subsection{Analyzing Features of Human \emph{versus} Generated Texts} \label{sec:case-study-tg}

\begin{tcolorbox}[colback=black!2!white,colframe=black!17!white,arc=0mm,fonttitle=\bfseries,coltitle=black,title=\icon~~Text generation]
 \emph{Carol is an NLP practitioner working on generative large language models (LLMs). She is interested in exploring the differences between texts written by humans and those generated by LLMs in terms of length, lexical diversity, and word use.}
\end{tcolorbox}

\noindent Carol uses the Human ChatGPT Comparison Corpus~\citep[HC3;][]{guo-etal-2023-hc3}, loading it from the \hficon~Hugging Face hub.\footnote{\url{https://huggingface.co/datasets/Hello-SimpleAI/HC3}.} HC3 comprises answers written by humans and ChatGPT-generated responses given the same questions across five domains. Through \variationist, Carol specifies two text columns of interest: \texttt{human\_answers} and \texttt{chatgpt\_answers}. She sets bigrams as units with lowercase normalization, and specifies stopword removal in English, further adding ``url'' and numbers from 0 to 9 as extra unigrams to remove. She defines \texttt{stats}, \texttt{root\_ttr}, and \texttt{npw\_pmi} as metrics in order to analyze different aspects of the texts. The other parameters are left with default values.

\begin{figure}[!ht]
\centering
\begin{subfigure}{.49\textwidth}
  \centering
  \includegraphics[width=.975\linewidth]{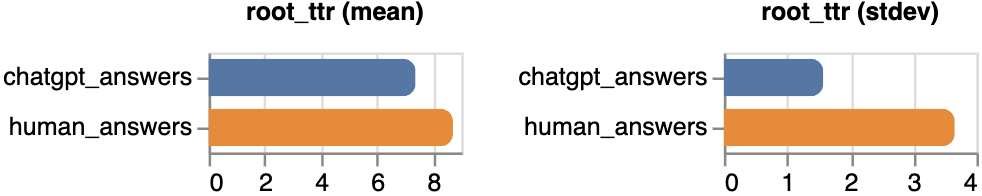}
  \caption{Bar charts comparing the lexical variety of human and ChatGPT-generated answers according to \texttt{root\_ttr}.}
  \label{fig:cs3a}
\end{subfigure} \par\bigskip
\begin{subfigure}{.49\textwidth}
  \centering
  \includegraphics[width=.975\linewidth]{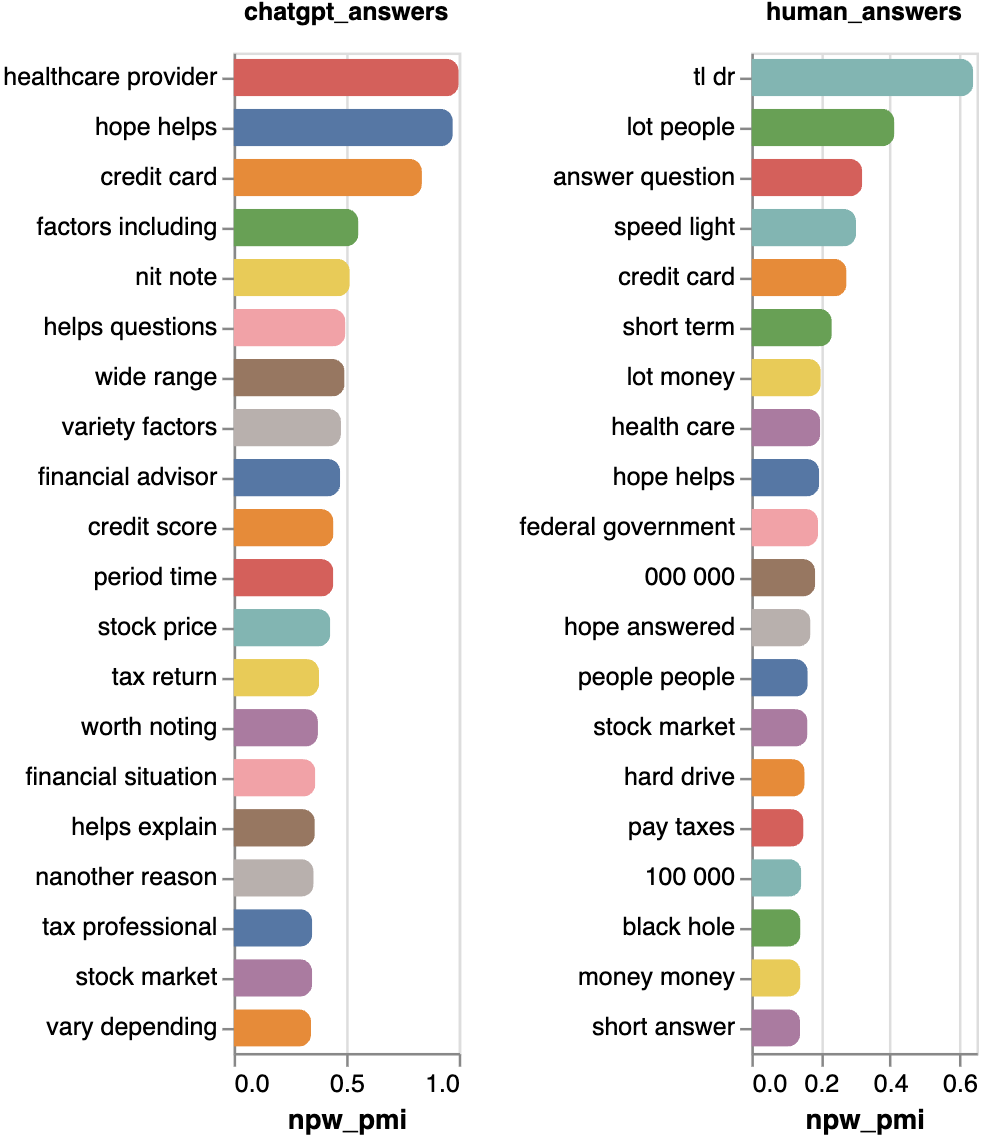}
  \caption{Bar charts showing the top-$k$ ($k$=20) informative bigrams for human and ChatGPT-generated answers according to \texttt{npw\_pmi}.}
  \label{fig:cs3b}
\end{subfigure}
\caption{Example visualizations for the text generation case study. The charts present some characteristics at the lexical level for human and ChatGPT-generated texts.}
\label{fig:cs3}
\end{figure}

By looking at the summary \texttt{stats}, Carol finds that human answers are on average much longer than ChatGPT-generated ones (i.e., 98.26 \emph{vs} 73.66 units) and that the vocabulary size of human answers is almost two times that of synthetic responses (i.e., 1.60M \emph{vs} 0.87M). Moreover, human-produced answers are more varied in terms of \texttt{root\_ttr}, also exhibiting a larger standard deviation compared to ChatGPT-generated ones (cf.~Figure~\ref{fig:cs3a}). Finally, by looking at the top-$k$ ($k$=20) bigrams associated to human and ChatGPT texts (Figure~\ref{fig:cs3b}), Carol finds that human answers appear to include terms that are more commonly used in everyday situations (e.g., ``lot people'', ``lot money''), while ChatGPT answers tend to include language that is more formal and less conversational, such as ``healthcare provider'' or ``variety factors''. In addition to this, it is clear from the \texttt{npw\_pmi} scores that the distribution of bigrams is a bit more balanced for human-authored texts, while ChatGPT appears to produce texts that include very specific bigrams with a much higher frequency. This might be a consequence of the different lexical variety between ChatGPT and human-authored texts.

As a future exploration, Carol aims to investigate which $n$ co-occurrences of language units appear to be strongly associated to synthetic responses within specific domains (e.g., finance). This can be done by providing the \texttt{source} column of the HC3 dataset as an additional variable to \variationist.

\subsection{Food for Thought} \label{sec:food-for-thought}

The potential applications of \variationist~are many. For instance, it can be used for advancing research on sociolects such as African-American English~\citep{blodgett-etal-2016-demographic}, to study linguistic reclamation and more generally investigate semantic change over time, or to conduct qualitative error analyses for model predictions (i.e., to unveil which language units are more informative of a wrong class). Moreover, it can be used to compare tokenizers and their effect on specific language varieties. We leave those areas to the reader as potential avenues for future applications of \variationist.

\section{Related Work}

There exist many tools for data exploration in literature, especially in the field of corpus linguistics (see~\citet{anthony-2013-critical} for an overview). However, there is currently a lack of a unified tool to serve diverse research communities that goes beyond descriptive statistics and basic charts, and that handles many variables at once in a simple fashion. The closest work to \variationist~is \hficon~Hugging Face's Data Measurements Tool~\citep[DMT;][]{luccioni-etal-2021-introducing}. However, it does not consider multiple texts and variables in the analysis, and it does not provide customization and flexibility in terms of units, metrics, tokenizers, and charts. \variationist~serves to fill this gap in literature.

\section{Conclusion}

We introduced \variationist, a modular, customizable, and easy-to-use analysis and visualization tool that aims at helping researchers in understanding language variation and unveiling potential biases in written language corpora across many dimensions. Through the case studies of \emph{Alice}, \emph{Bob}, and \emph{Carol}, we showed the potential of our tool in answering different research questions across disciplines. We hope that our work will also be useful to \emph{Dave}, a fourth fictional character who unfortunately looks at the data very rarely before using it, to begin to reconsider the pivotal role of exploring data before using it for training language models.

\section*{Ethics and Broader Impact Statement}

\variationist~serves as a tool to support the research community in better understanding the diversity in language use and unveiling quality issues and harmful biases in textual data. As a result, we do not foresee specific ethical concerns related to our work and, on the contrary, we hope that \variationist~will give additional means to explore datasets and raise awareness among researchers on the paramount importance of looking at the data.

\variationist~has been designed following an inclusion-first approach, i.e., by avoiding common language-specific assumptions to better support its application across many language varieties. As a limitation, we acknowledge that \variationist~is currently limited to the lexical level on written data. We aim to extend its functionalities to also cover other linguistic aspects such as grammar as well as the speech modality in the next releases.

\section*{Acknowledgments}
This work has been funded within the European Union’s ISF program under grant agreement No. 101100539 (PRECRISIS) and by the European Union’s Horizon Europe research and innovation program under grant agreement No.~101070190 (AI4Trust).

\bibliography{custom}

\appendix

\section*{Appendix} \label{sec:appendix}

\section{\variationist's Resources} \label{sec:resources}

All the resources related to \variationist~are made publicly-available to the research community. Table~\ref{tab:materials} lists all the links to these resources.

\begin{table}[h]
  \centering
  \resizebox{1.0\linewidth}{!}{%
  \begin{tabular}{lp{7.5cm}}
    \toprule
    \textbf{Resource} & \textbf{URL} \\
    \midrule
    Code & \url{https://github.com/dhfbk/variationist} \\[0.05cm]
    Library & \url{https://pypi.org/project/variationist} \\[0.05cm]
    Docs & \url{https://variationist.readthedocs.io} \\[0.05cm]
    Video & \url{https://github.com/dhfbk/variationist} \\[0.05cm]
    Tutorials & \url{https://github.com/dhfbk/variationist/tree/main/examples} \\
    \bottomrule
  \end{tabular}
  }%
  \caption{\label{tab:materials} Publicly-available \variationist~resources.}
\end{table}

\section{Credits}

Emojis are from the Google Noto Emoji set (\url{https://github.com/googlefonts/noto-emoji}).

\end{document}